%% file: emnlp2023.tex
\pdfoutput=1

\documentclass[11pt]{article}

\usepackage[]{EMNLP2023}

\usepackage{times}
\usepackage{latexsym}
\usepackage{graphicx}
\usepackage{tabularx}
\usepackage{multirow}
\usepackage{mathtools}
\usepackage{enumitem,kantlipsum}
\usepackage[export]{adjustbox}
\usepackage{booktabs}
\usepackage{caption}
\usepackage{subcaption}
\usepackage{tikz}
\usepackage{cleveref}
\usepackage[linesnumbered,ruled,vlined]{algorithm2e}
\usepackage{hyperref}

\usepackage[T1]{fontenc}

\usepackage[utf8]{inputenc}

\usepackage{microtype}

\usepackage{inconsolata}
\newcommand{\mailto}[2]{\texttt{\href{mailto:#1}{#2}}}

\newcommand{\example}[1]{\emph{``#1''}}
\newcommand\tool{\textsc{Champ}}

\title{\tool{}: Efficient Annotation and Consolidation of Cluster Hierarchies}

\author{Arie Cattan\textsuperscript{1} \quad
        Tom Hope\textsuperscript{2,3} \quad 
        Doug Downey\textsuperscript{2} \quad 
        Roy Bar-Haim\textsuperscript{4} \\
        \textbf{Lilach Eden\textsuperscript{4}} \quad
        \textbf{Yoav Kantor\textsuperscript{4}} \quad
        \textbf{Ido Dagan\textsuperscript{1}} \\
        \textsuperscript{1}Computer Science Department, Bar Ilan University  \\
        \textsuperscript{2}Allen Institute for Artificial Intelligence  \\
        \textsuperscript{3}School of Computer Science, The Hebrew University of Jerusalem   \\
        \textsuperscript{4}IBM Research       \\ 
  {
  \mailto{arie.cattan@gmail.com}{\texttt{arie.cattan@gmail.com}} 
  } 
 }

\begin{document}
\maketitle

\input{sections/00_abstract}
\input{sections/01_intro}

\input{sections/02_bg}

\input{sections/03_chat}

\input{sections/04_application}
\input{sections/05_release}
\input{sections/06_conclusion}

\section*{Acknowledgements}

This work was supported by the Israel Science Foundation (grant no. 2827/21). Arie Cattan is partially supported by the PBC fellowship for outstanding PhD candidates in data science.

\bibliography{anthology,custom}
\bibliographystyle{acl_natbib}

\appendix

\section{Appendix}
\label{sec:appendix}

\begin{figure*}[t]
     \centering
     \includegraphics[width=0.9\textwidth, frame]{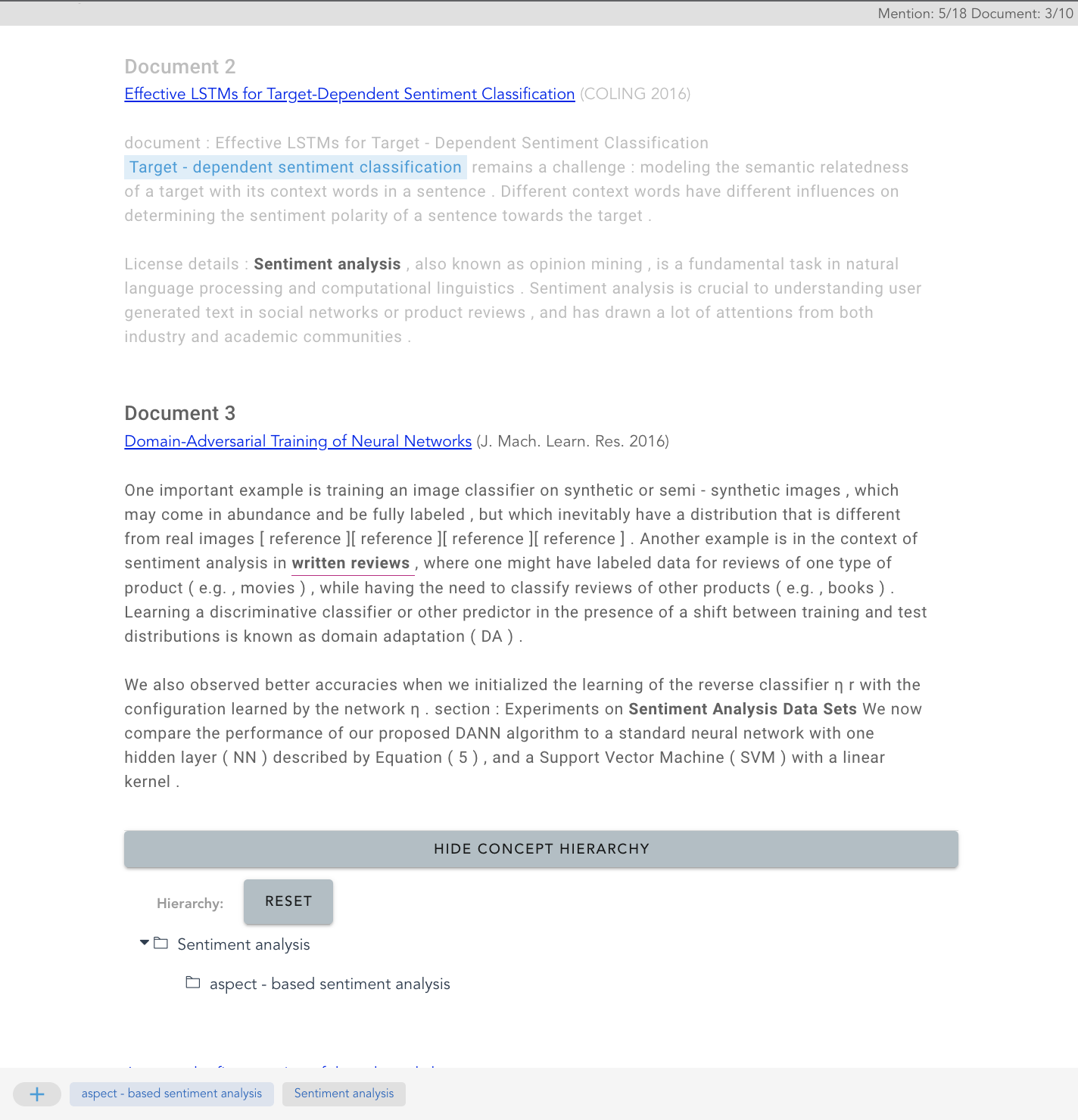}
     \caption{User interface for annotating hierarchy of clusters over textual spans that appear within surrounding context.}
     \label{fig:scico}
\end{figure*}

Figure~\ref{fig:scico} shows the interface of \tool{} for annotating a hierarchy of clusters over text spans appearing in their context. This example was taken from \textsc{SciCo}.

\end{document}

%% file: sections/00_abstract.tex
\begin{abstract}

Various NLP tasks require a complex hierarchical structure over nodes, where each node is a cluster of items. Examples include generating entailment graphs, hierarchical cross-document coreference resolution, annotating event and subevent relations, etc. To enable efficient annotation of such hierarchical structures, we release \tool{}, an open source tool
allowing to incrementally construct both clusters and hierarchy simultaneously over any type of texts. This incremental approach significantly reduces annotation time compared to the common pairwise annotation approach and also guarantees maintaining transitivity at the cluster and hierarchy levels. Furthermore, \tool{} includes a consolidation mode, where an adjudicator can easily compare multiple cluster hierarchy annotations and resolve disagreements. 

\vspace{1em} 
\hspace{.5em}\includegraphics[width=1.25em,height=1.25em]{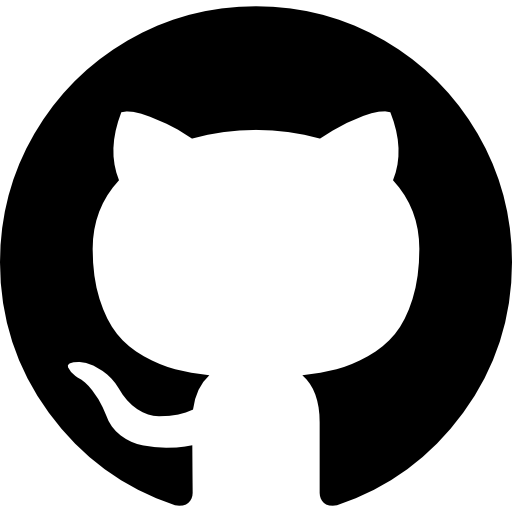}\hspace{.5em}\parbox{\dimexpr\linewidth-2\fboxsep-2\fboxrule}{\url{https://github.com/ariecattan/champ}}



\end{abstract}

%% file: sections/01_intro.tex
\section{Introduction}


\begin{figure}
    \centering
    \includegraphics[scale=0.7]{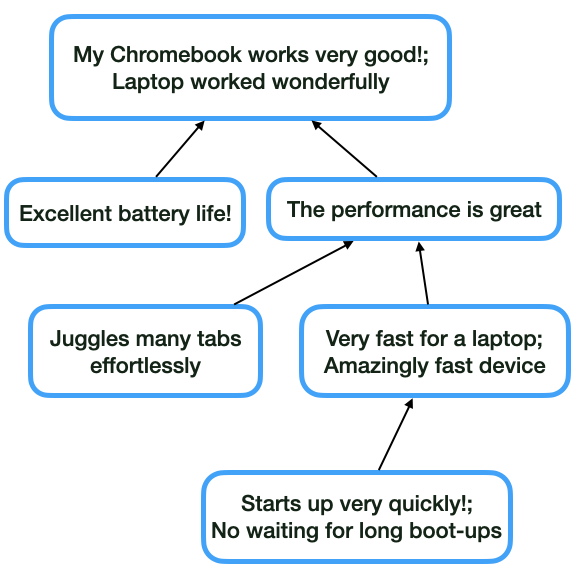}
    \caption{Example of hierarchy of clusters from \textsc{ThinkP}~\citep{cattan-etal-2023-key}. Nodes group similar statements together and arrows represent child-parent relations, relating specific statements to more general ones.}
    \label{fig:kph}
\end{figure}

In numerous annotation tasks, the annotator needs to perform individual and independent decisions. Such tasks include Named Entity Recognition (NER), text categorization and part-of-speech tagging, among others~\citep{stenetorp-etal-2012-brat, yimam-etal-2013-webanno, samih-etal-2016-sawt, yang-etal-2018-yedda, tratz-phan-2018-web, mayhew-roth-2018-talen}. However, certain annotation tasks are more demanding because they involve the construction of a complex structure that must satisfy global constraints. One such complex structure is clustering, where annotated clusters must respect the equivalence relation. Specifically, if items A and B belong to the same cluster, and items B and C also belong to the same cluster, then A and C must belong to the same cluster as well. Another prominent example of a global structure is hierarchy, where typically, if A is an ancestor of B and B is an ancestor of C, then A must also be an ancestor of C. 



In this work, we focus on annotating a \textit{hierarchy of clusters}, a global structure that combines the constraints of both clustering and hierarchy, thereby posing further challenges. In this hierarchy, nodes are clusters of (text) items, where each node can have at most a single parent, as illustrated in Figure~\ref{fig:kph}. Annotating a hierarchy of clusters is relevant for a multitude of tasks, such as hierarchical cross-document coreference resolution~\citep{cattan2021scico}, structured summarization as a hierarchy of key points~\citep{cattan-etal-2023-key}, entailment graph construction~\citep{berant-etal-2012-efficient} and event-subevent relations detection~\citep{ogorman-etal-2016-richer, wang-etal-2022-maven}. While there are some annotation tools for annotating either clustering or a hierarchy~(§\ref{subsec:bg_tools}), to the best of our knowledge there is no available tool allowing to annotate a hierarchy of clusters simultaneously within the same tool.

To address this need, we introduce \tool{} (\textbf{C}luster \textbf{H}ierarchy \textbf{A}nnotation for \textbf{M}ultiple \textbf{P}articipants), an intuitive and efficient tool for annotating a hierarchy of clusters in a globally consistent manner, supporting multiple annotators~(§\ref{sec:tool}). Specifically, annotators are presented with input text spans one by one and form \textit{incrementally} and \textit{simultaneously} the clusters and their hierarchy~(§\ref{subsec:hierarchy}). 


Additionally to the annotation process, we develop an adjudication mode for easily comparing multiple annotated hierarchies of clusters~(§\ref{subsec:adjudication}). This mode can be used either by an adjudicator, which is typically a more reliable annotator, or by the original annotators during discussions to resolve conflicts. Indeed, adjudication is crucial to ensure quality in general~\citep{roit-etal-2020-controlled, klein-etal-2020-qanom}, and particularly important for our structure, requiring a more challenging global annotation. 




We demonstrate the use of \tool{} in two notably different use-cases, both involving annotating hierarchies of clusters: hierarchical cross-document coreference resolution~\citep{cattan2021scico} and key point hierarchy~\citep{cattan-etal-2023-key}. In both settings, \tool{} is significantly more efficient than a pairwise annotation approach, in which the relation between each pair of items is annotated independently. Moreover, our consolidation phase enhances the annotation quality, yielding an improvement of 5-6 F1 points~\citep{cattan-etal-2023-key}.


\tool{} was implemented on top of \textsc{CoRefi}~\cite{bornstein-etal-2020-corefi}, which was initially designed for coreference, and allowed only standard (non-hierarchical) annotation. \tool{} includes a WebComponent, which can easily be embedded into any HTML page, including popular crowdsourcing platforms such as Amazon Mechanical Turk. We also develop an annotation portal (the link appears in our github repository), allowing users to perform online the annotation task and dataset developers to effortlessly compute inter-annotator agreement.

Overall, \tool{} is an intuitive tool for efficiently annotating and adjudicating hierarchies of clusters. We believe that \tool{} will remove barriers when annotating such challenging global tasks and will facilitate future dataset creation.


%% file: sections/02_bg.tex
\section{Background}
\label{sec:bg}


\subsection{Tools for Annotating Global Structures}
\label{subsec:bg_tools}

Certain NLP tasks involve a structure that should be annotated in a global manner due to mutually dependent labels. In this work, we focus on two specific structures: clustering and hierarchy.

A prominent clustering task is coreference resolution, where the goal is to group mention spans into clusters. This implies 
that if A and B are coreferent and B and C are coreferent, then A and C should also be coreferent. However, early tools for coreference annotation relied on a series of local binary decisions over all possible mention pairs~\citep{stenetorp-etal-2012-brat, 10.1145/2361354.2361394, landragin-etal-2012-analec, kopec-2014-mmax2, chamberlain-etal-2016-phrase}. In contrast, cluster-based tools aim for global annotation by directly assigning mentions to clusters~\citep{ogren-2006-knowtator, girardi-etal-2014-cromer, Reiter2018ag, oberle-2018-sacr, aralikatte-sogaard-2020-model, bornstein-etal-2020-corefi, gupta-etal-2023-ezcoref}. Among these cluster-based tools, \textsc{CoRefi}~\citep{bornstein-etal-2020-corefi} stands out for its beneficial features that enable cost-effective and efficient annotation. These features include quick keyboard operations (instead of slow drag-and-drop), an onboarding mode for training annotators on the task, and a reviewing mode that facilitates systematic review and quality improvement of a given annotation (as described in §\ref{subsec:bg_cons}).

Some other tasks such as taxonomy induction and entailment graph construction also involve structures (e.g., graphs, DAG, hierarchy) that impose global transitivity constraints. For example, if a taxonomy includes the relationships ``A is a kind of B'' and ``B is a kind of C'', then it follows that A must also be a kind of C. Yet, for example,~\citet{berant-etal-2011-global} annotated an entailment graph dataset by annotating all possible edges between predicates, resulting in a complexity of $\mathcal{O}(n^2)$. Subsequent works follow the pairwise approach but apply some heuristics for reducing the number of annotations~\citep{levy-etal-2014-focused, Kotlerman2015TextualEG}.
Closely related to taxonomy, the Redcoat annotation tool~\citep{stewart-etal-2019-redcoat} allows to annotate hierarchical entity typing, while allowing to modify the hierarchy during annotation.

To the best of our knowledge, there is no available tool that supports joint annotation of a hierarchy of clusters, as proposed in \tool{}.



\subsection{Consolidation of Multiple Annotations}
\label{subsec:bg_cons}

To promote quality, datasets often rely on multiple annotators per instance, especially when the annotation is obtained via crowdsourcing. Then, the annotations can be combined either \textit{automatically}, using simple majority vote or more sophisticated aggregation techniques~\citep{Dawid1979MaximumLE,  Raykar2010LearningFC, hovy-etal-2013-learning, passonneau-carpenter-2014-benefits, paun-etal-2018-comparing}, or \textit{manually}, by asking the annotators themselves or a more reliable annotator to adjudicate and resolve annotation disagreements~\citep{pradhan-etal-2012-conll, roit-etal-2020-controlled, pyatkin-etal-2020-qadiscourse, klein-etal-2020-qanom}. However, those aggregation methods were mostly investigated for classification tasks where each instance can be annotated independently, but not for global tasks, like those discussed above~(§\ref{subsec:bg_tools}). 




To the best of our knowledge, \textsc{CoRefi}~\citep{bornstein-etal-2020-corefi} is the only annotation tool that supports \textit{manual} reviewing of a global structure annotation, specifically for coreference annotation. In this interface, the reviewer is shown the annotated mentions one by one along with the original annotator's cluster assignment.
The reviewer can then decide whether to retain the original annotation or to make a different clustering assignment. However, showing the original cluster assignment of each mention in turn is not straightforward, because earlier reviewer decisions may have deviated from the original clustering annotation. For instance, consider a scenario where the original annotator creates a cluster with the mentions ${x, y, z}$. Subsequently, the reviewer decides that $y$ should not be linked to $x$ but should instead form a new cluster. At this point, when the reviewer encounters the mention $z$, it becomes uncertain whether it should be considered by the original annotation as linked with ${x}$ or ${y}$. To address this issue, when the reviewer is shown a mention $m$, the candidate clusters implied by the original annotation becomes the \textit{set} of clusters in the current reviewer's clustering configuration that include at least one of the previously annotated antecedents of $m$ according to the original annotation.

While the reviewing mode in \textsc{CoRefi} is effective, an important limitation is that it enables reviewing only a single annotation, not supporting the consolidation of multiple annotations, as common in NLP annotation setups. We address this need in \tool{} by supporting consolidation of multiple annotations~(§\ref{subsec:adjudication}).



%% file: sections/03_chat.tex
\section{\tool{}}
\label{sec:tool}

We present \tool{}, a new tool for annotating a \textit{hierarchy of clusters}.
To annotate such a structure, the annotators are provided with a list of input spans, denoted as $S=\{s_1, ..., s_n\}$, that they need to group into disjoint clusters of semantically equivalent spans $\mathcal{C} = \{C_1, ..., C_k\}$. In addition, annotators need to form a \textit{directed} forest $G = (\mathcal{C}, E)$, constituting a Directed Acyclic Graph (DAG) in which every node---representing the cluster $C_i$---has no more than one parent. Within this structure, each edge $e_{ij}$ represents a hierarchical relation between clusters $C_i \xrightarrow{} C_j$, signifying that $C_i$ is a child of $C_j$. Considering the example in Figure~\ref{fig:kph}, the cluster \emph{\{Starts up very quickly, No waiting for long boot-ups\}} is more specific than the cluster \emph{\{Very fast for a laptop, Amazingly fast device\}}. Importantly, input spans can be standalone spans (as in Figure~\ref{fig:kph}) or appear within a surrounding context. For the remainder of this section, we will focus on demonstrating \tool{} using standalone spans, while an example featuring spans within context is provided in Appendix~\ref{sec:appendix}.

We next describe the core annotation interface~(§\ref{subsec:hierarchy}), and then present the \textit{adjudication} mode, which allows to effectively compare multiple annotations and build a consolidated hierarchy of clusters~(§\ref{subsec:adjudication}). 

\subsection{Cluster Hierarchy Annotation}
\label{subsec:hierarchy}

\begin{figure*}[t]
     \centering
     \includegraphics[width=0.8\textwidth, frame]{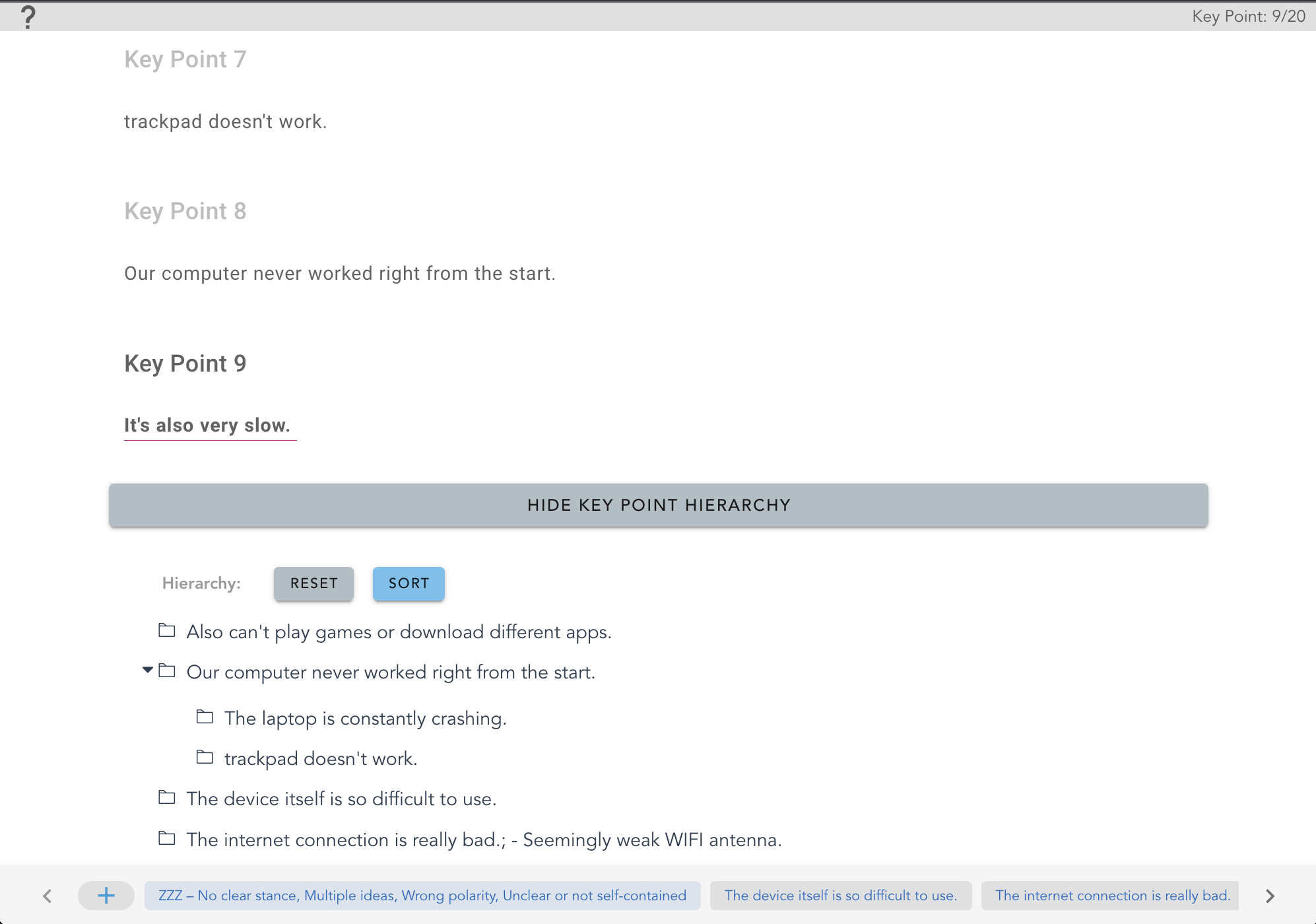}
     \caption{User interface for annotating both clustering and hierarchical relations between clusters. The current statement to assign is underlined in purple: ``It's also very slow''. The annotator can decide whether to add it to an existing cluster, in which case it will be concatenated in the display of the corresponding node in the hierarchy, separated by ``;'', or to open a new cluster, in which case a new node will be automatically added to the hierarchy, initiated under the root.}
     \label{fig:hierarchy}
\end{figure*}

Figure~\ref{fig:hierarchy} shows the annotation interface in \tool{}.

A naive approach for supporting the annotation of a hierarchy of clusters would involve two separate steps: (1) cluster input spans and (2) construct a hierarchy over the fixed annotated clusters. Although straightforward, this method lacks the flexibility for annotators to modify the clustering annotation while simultaneously working on the hierarchy. This inflexibility is problematic since typically many annotation decisions fall at the intersection of clustering, which reflects semantic equivalences, and hierarchy, which denotes the relationships between more general and specific clusters (e.g., \emph{Takes a long time for check in} vs. \emph{The absolute worst check in process anywhere}). Moreover, employing two separate annotation steps would burden annotators with the additional challenge of remembering the context of each cluster during hierarchy annotation.

Therefore, we propose an \textit{incremental} approach for annotating both the clustering and the cluster hierarchy together as a single annotation task, which we develop upon \textsc{CoRefi}~\citep{bornstein-etal-2020-corefi}. At initialization, the first span is automatically assigned to the first cluster $C_1$ and to a corresponding node in the hierarchy. Then, for each subsequent span $s$, the annotator first decides its cluster assignment, by choosing whether to assign $s$ to an existing or a new cluster. In the latter case, a new node is automatically created in the hierarchy under the root and the annotator can drag it to its right position in the current hierarchy. 
Considering the example in Figure~\ref{fig:hierarchy}, the current span to annotate $s$ is ``It's also very slow'' (underlined in purple), the current clusters $\mathcal{C}$ are shown in the cluster bank (in the footer of the screen), and the current hierarchy is shown in the lower portion of the window.

Importantly, when the annotator re-assigns a previously assigned span to another cluster, \tool{} will automatically update nodes and relations in the hierarchy. Keeping in sync cluster assignments and hierarchy is not trivial because different clustering modifications will have different effects on the resulting hierarchy. In particular, we consider the following cases of re-assigning the span $s$:

\begin{enumerate}[wide, labelwidth=!, labelindent=0pt]
    \item From a \textit{singleton} cluster $C_i$ to a cluster $C_j$: $s$ will be added to $C_j$ and $C_i$'s children will move under $C_j$.
    \item From a \textit{non-singleton} cluster $C_i$ to a cluster $C_j$: $s$ will be added to $C_j$ but $C_i$'s children will stay under $C_i$.  
    \item From a cluster $C_i$ to a new singleton cluster: a new node $C_j$ will be created in the hierarchy and will be initially situated as a sibling of $C_i$.\footnote{We take this approach because, when annotators re-assign $s$ to a standalone cluster, their intention is not to eliminate the hierarchical relationship between $s$ and its parent cluster.} Annotators can then drag it to its desired place.
\end{enumerate}

This hierarchy update procedure is a key ingredient for enabling the annotation of hierarchy of clusters as a single task.




\subsection{Adjudication}
\label{subsec:adjudication}

\begin{figure*}
\centering
\begin{subfigure}[t]{.47\textwidth}
  \centering
  \includegraphics[width=.98\linewidth, frame]{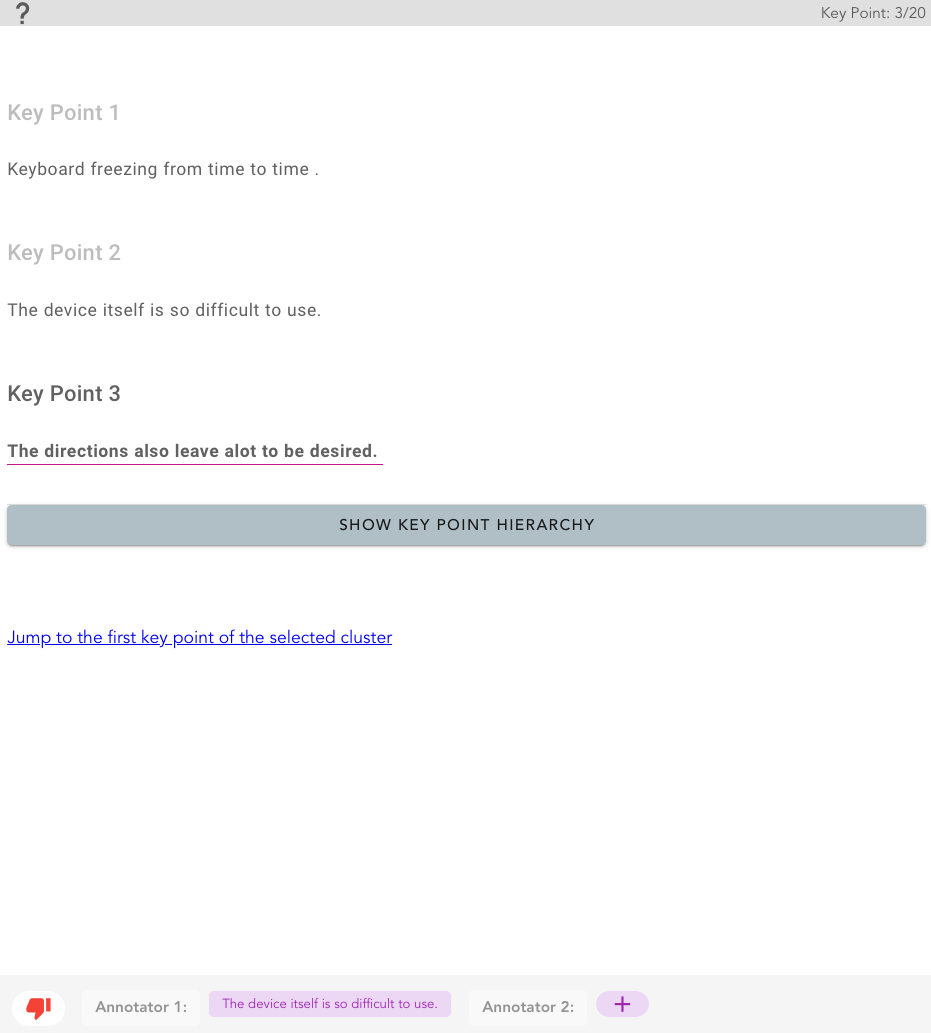}
  \caption{Clustering consolidation. The thumb-down at the bottom left of the screen indicates a clustering disagreement between the annotators for the span \example{The directions also leave a lot to be desired}. Annotator A1 assigned it to \example{The device itself is so difficult to use} while annotator A2 created a new cluster, as indicated in purple.}
  \label{fig:cons1}
\end{subfigure}%
\quad
\begin{subfigure}[t]{.47\textwidth}
  \centering
  \includegraphics[width=.98\linewidth, frame]{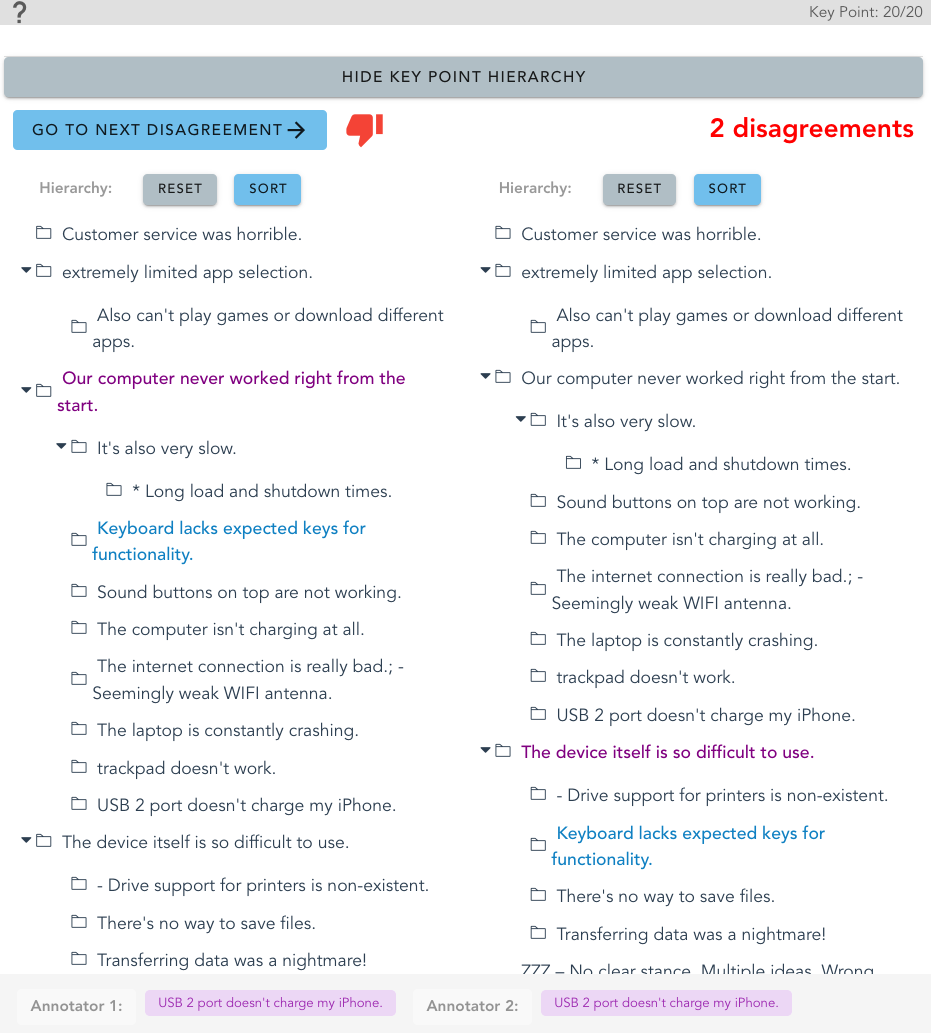}
  \caption{Hierarchy consolidation. The red thumb-down near the ``Go to next disagreement'' button indicates a hierarchy disagreement for the node \example{Keyboard lacks expected keys for functionality}. Annotator A1 placed it under \example{Our computer never worked right from the start}, while A2 placed it under \example{The device itself is so difficult to use.}}
  \label{fig:hierarchy_cons}
\end{subfigure}
\caption{Adjudication of multiple annotations of hierarchy of clusters.}
\label{fig:adjudication}
\end{figure*}

In order to facilitate the manual adjudication of multiple hierarchy annotations by different annotators, we added an adjudication mode within \tool{} that supports easily identification and resolution of disagreements between any number of annotations. This mode can be used by an adjudicator, which is usually a more reliable annotator, or by the original annotators during discussions to resolve conflicts.

Comparing multiple annotations of a hierarchy of clusters can be challenging due to variations in annotators' clustering assignments, leading to different sets of nodes in the respective hierarchies. To illustrate this issue, consider a scenario where annotator A1 annotates the relation $\{s_1, s_2, s_4\} \xrightarrow{} \{s_3, s_5\}$, while A2 annotates $\{s_1, s_2, s_6\}\xrightarrow{} \{s_3\}$ and $\{s_4\} \xrightarrow{} \{s_5\}$.
The two hierarchies have similarities (e.g. both cluster $s_1$ and $s_2$ together and have $s_5$ as a parent of $s_4$) but differ in other ways, 
making their 
adjudication process non-trivial.

To tackle this problem, we decoupled the adjudication process into two consecutive stages, adjudicating separately clustering and hierarchy decisions, as illustrated in Figure~\ref{fig:adjudication}. 


In the first step, the adjudicator is shown the annotated spans in a sequential manner, along with the cluster assignments of each of the original annotations. To achieve this, we leverage the reviewing procedure that \textsc{CoRefi} applies for reviewing a single clustering annotation~(§\ref{subsec:bg_cons}), implement it separately to each original annotation. We then present to the adjudicator a set of candidate clusters per original annotation. These sets of candidates are displayed in purple at the bottom of the screen, as illustrated in Figure~\ref{fig:cons1}.

It should be pointed out here that resolving a cluster assignment disagreement means that the adjudicator alters the assignment for at least one of the annotators. Therefore, we apply the hierarchy update procedure~(§\ref{subsec:hierarchy}) to the modified annotations, in order to update accordingly the involved cluster nodes and their hierarchical relations. Considering the example in Figure~\ref{fig:cons1} with a clustering disagreement for the span \example{The directions also leave a lot to be desired ($s_1$)}. In this instance, annotator A1 has merged it with \example{The device itself is so difficult to use ($s_2$)}, while annotator A2 has designated it as a singleton cluster in the hierarchy, as highlighted by the purple `+' button.
If the adjudicator follows A1's decision, A2's hierarchy will be restructured to combine spans $\{s_1, s_2\}$ into the same cluster. Conversely, siding with A2's decision will separate $s_2$ from $s_1$ in A1's hierarchy. This automatic process ensures that the modified hierarchies will include the exact same set of nodes (clusters) $\mathcal{C}$ at the end of the clustering consolidation step.

In the second step of hierarchy adjudication, as the sets of nodes $\mathcal{C}$ in the hierarchies of all annotators are identical, a disagreement arises when a node $C_i \in \mathcal{C}$ has a different direct parent in different hierarchies. To efficiently identify such discrepancies, the adjudicator can click on the ``Go To Next Disagreement'' button, which highlights the node $C_i$ in blue along with its direct parent in violet on all input hierarchies. As shown in Figure~\ref{fig:hierarchy_cons}, for instance, the node \example{Keyboard lacks expected keys for functionality} was placed under \example{Our computer never worked right from the start} by A1, and under \example{The device itself is so difficult to use} by A2. The adjudicator then decides the correct hierarchical relation, manually updates the other hierarchies accordingly, and moves on to the next disagreement. Once all hierarchical disagreements have been resolved, the adjudicator can confidently submit the obtained consolidated hierarchy.


%% file: sections/04_application.tex
\section{Applications}
\label{sec:application}

We used \tool{} for annotating datasets for two different tasks that require annotating of hierarchy of clusters:

\begin{enumerate}[wide, labelwidth=!, labelindent=0pt]
    \item SciCo~\citep{cattan2021scico}, a dataset for the task of hierarchical cross-document coreference resolution (H-CDCR). In this dataset, the inputs are paragraphs from computer science papers with highlighted mentions of scientific concepts, specifically mentions of tasks and methods. The goal is to first cluster all mentions that refer to the same concept (e.g., \textit{categorical image generation} $\xleftrightarrow{}$ \textit{class-conditional image synthesis}) and then infer the referential hierarchy between the clusters (e.g., \textit{categorical image generation} $\xrightarrow{}$ \textit{image synthesis}).

    \item \textsc{ThinkP}~\citep{cattan-etal-2023-key}, a recent benchmark of key point hierarchies, where each key point is a concise statement relating to a particular topic~\citep{bar-haim-etal-2020-arguments}. Key point hierarchies were proposed as a novel structured representation for large scale opinion summarization. The nodes in these graphs group statements conveying the same opinion (e.g., \textit{the cleaning crew is great!} $\xleftrightarrow{}$ \textit{housekeeping is fantastic}) while the edges indicate hierarchical specification-generalization relationships between nodes (e.g., \textit{housekeeping is fantastic} $\xrightarrow{}$ \textit{the personnel is great}). The entailment graphs in \textsc{ThinkP} are designed in a hierarchical form, where each node has at most a single parent. 
    
\end{enumerate}

Despite the different nature of these tasks and their unit of annotation (i.e., standalone state,emts vs. concept spans in context), we seamlessly leveraged \tool{} for both with minimal effort (using a simple JSON configuration schema), as both tasks involve annotating a hierarchy of clusters.


In our experiments, we observed that annotating or consolidating a hierarchy of clusters for fifty statements takes approximately one hour~\citep{cattan-etal-2023-key}. In contrast, collecting annotations for all possible pairs, as commonly done in prior datasets for entailment graphs~\citep{berant-etal-2011-global}, would have been much more expensive since it would require at least 1225 decisions on average for our data, which would obviously take much more than one hour. Furthermore, unlike the pairwise annotation approach, our incremental method for constructing a hierarchy of clusters guarantees that the resulting annotation will respect the global constraint of transitivity. Finally, our experiments also revealed that the consolidation mode significantly enhances human performance, yielding a gain of 5-6 F1 points~\citep{cattan-etal-2023-key}.



%% file: sections/05_release.tex
\section{Implementation Details and Release}

We implement \tool{} on top of \textsc{CoRefi}~\cite{bornstein-etal-2020-corefi}, using the \url{Vue.js} framework, that we open source under the permissive MIT License. Following \textsc{CoRefi}, we release \tool{} as a WebComponent, which can easily be embedded into any HTML page, including popular crowdsourcing platforms such as Amazon Mechanical Turk. Both the annotation and consolidation processes share the same interface and are easily configurable using a straightforward JSON schema. We also develop an annotation portal where users can upload a configuration file (either for annotation or adjudication), perform the annotation task and download it upon completion. This portal also provides the capability to upload multiple annotation files from various annotators and to compute the inter-annotator agreement. As such, \tool{} is not only easy-to-use for annotators, but it is also easy to setup and manage for dataset developers. 

%% file: sections/06_conclusion.tex
\section{Conclusion}

This paper aims to foster research on global annotation tasks by introducing \tool{}, an efficient tool designed for annotating a \textit{hierarchy of clusters}. This annotation tool also incorporates an adjudication mode that conveniently supports  identification and consolidation of annotators' disagreements. As \tool{} enables efficient and high-quality annotation, we believe that it will facilitate the creation of datasets for various tasks involving this complex structure, and will inspire tool development for other global annotation tasks. 
